\documentclass[11pt]{article}

\usepackage[preprint]{acl}

\usepackage{times}
\usepackage{latexsym}
\usepackage[T1]{fontenc}
\usepackage[utf8]{inputenc}
\usepackage{microtype}
\usepackage{inconsolata}
\usepackage{graphicx}
\graphicspath{{figures/}{Figures/}{latex/figures/}{latex/Figures/}}
\usepackage{booktabs}
\usepackage{multirow}
\usepackage{amsmath}
\usepackage{amssymb}
\usepackage{xcolor}
\usepackage{colortbl}
\usepackage{enumitem}
\usepackage[most]{tcolorbox}

\definecolor{lightgray}{gray}{0.92}
\definecolor{lightblue}{rgb}{0.90,0.95,1.00}
\definecolor{lightyellow}{rgb}{1.00,0.98,0.88}
\definecolor{asyellow}{rgb}{0.99,0.93,0.70}

\title{What Should Agents Say? Action-state Communication for \\ Efficient Multi-Agent Systems}

\author{
  Chen Huang,
  Yuhao Wu,
  Wenxuan Zhang
  \\
  Singapore University of Technology and Design \\
  \texttt{\{chen\_huang,yuhao\_wu\}@mymail.sutd.edu.sg},
  \texttt{wxzhang@sutd.edu.sg}
}

\begin{document}
\maketitle

\begin{abstract}
Multi-agent systems (MAS) built on large language models are typically organized around roles, pipelines, and turn schedules, while the content that agents pass to one another is often left as unconstrained natural language. 
However, this free-form communication can rapidly inflate token usage, consume the shared context window, and ultimately affect both system performance and inference cost.
We analyze five common inter-agent communication strategies across two MAS topologies, finding that no fixed strategy is universally optimal. Instead, effective inter-agent messages consistently preserve action-centered information needed by downstream agents.
Building on this, we propose the \textbf{PACT} (\textbf{P}rotocolized \textbf{A}ction-state \textbf{C}ommunication and \textbf{T}ransmission), which treats inter-agent communication
as a public state-update problem and projects each raw agent output into a compact action-state record before it enters shared history.
Across different MAS topologies, PACT consistently improves the performance--cost trade-off, achieving comparable or stronger task performance with substantially fewer tokens.
The gains extend to production coding harnesses: PACT lifts OpenHands' resolve
rate at $-10\%$ tokens-per-resolved, and is resolve-neutral on
SWE-agent while halving input tokens.
Our code is publicly available at \url{https://github.com/iNLP-Lab/PACT}.
\end{abstract}

\section{Introduction}
\label{sec:intro}

\begin{figure}[t]
  \centering
  \includegraphics[width=\columnwidth]{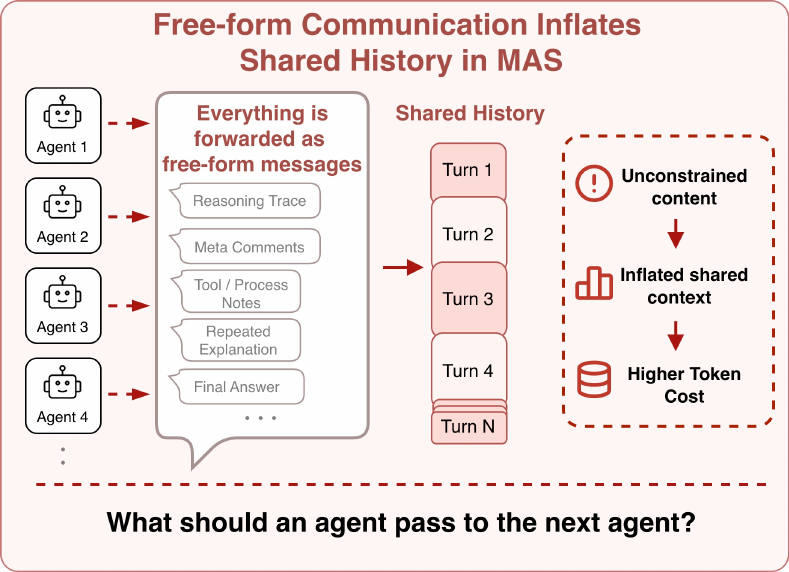}
  \caption{Free-form inter-agent messages accumulate in the shared history, forcing downstream agents to process unconstrained context and increasing token cost.
  }
  \label{fig:intro}
  \vspace{-10pt}
\end{figure}

Large language models (LLMs) are increasingly deployed as multi-agent
systems (MAS)~\citep{hong2024metagpt, cemri2026multi}. In these systems,
several agents collaborate to solve complex tasks that a single
model call handles poorly~\citep{wu2023autogen,chen2024agentverse,li2023camel}.
Such collaboration can take many forms, including debate~\citep{du2023improving,li2024improving}, sequential pipelines~\citep{zhang2024chain, zhao2025connecting,
zou2025latent}, retrieval-based relays~\citep{chen2025optima}, and
tool-using agents~\citep{yao2023react}.
The paradigm is already widely deployed in agentic coding assistants~\citep{anthropic_claude_code_2026, openai_codex_2026} and open agent platforms~\citep{wang2025openhands,yang2024sweagent}, decomposing a task across many cooperating model calls and outperform single-shot prompting. This effectiveness carries a cost: a MAS issues many model calls whose outputs feed one another, so it consumes far more tokens, often many times more, than a single model
solving the same task~\citep{bai2026ai, yu2026learning}.

This issue becomes especially severe when modern reasoning models are used in multi-turn settings~\citep{anthropic_claude_opus_4_7_2026, openai_gpt_5_5_2026, team2026kimi}: once an agent forwards a long internal reasoning trace, every downstream agent must repeatedly process it as part of the shared dialogue history~\citep{ramesh2025communicating}. 
As shown in Figure~\ref{fig:intro}, verbose or noisy messages are repeatedly reprocessed by downstream agents, causing token costs to compound across turns and potentially exhausting the context window before the task is completed, which can ultimately degrade performance~\citep{li2024improving, chen2025optima}.
However, existing MAS research has largely focused on roles, turn-taking schedules, and tool allocation~\citep{wu2023autogen,chen2024agentverse,wang2024mixture, qian2024chatdev}, while leaving the inter-agent message itself as mostly unconstrained free-form natural language. 
We argue that the inter-agent message is a central design lever: \textbf{the content of messages between agents directly shapes both MAS performance and token cost}.

Specifically, we ask: \emph{what should an agent send to another agent in a MAS}? 
We conduct a diagnostic analysis of inter-agent communication across two different MAS settings: a split-evidence interaction and a sequential pipeline. The analysis compares five common strategies: forwarding the agent's full free-form output, using native concise generation, keeping only the final conclusion, producing a short summary of the turn, and passing only the role artifact (\S\ref{sec:prelim}). 
The analysis shows that no single strategy is uniformly best across topologies: forwarding the full output is expensive and redundant, generic shortening can remove task-relevant information, conclusion-only messages are too lossy when the receiver lacks surrounding state, and artifact-only messages capture useful task content but still leave the receiver to infer the sender's intended action and grounding evidence. 
We therefore view inter-agent communication as a public state-update problem: the message should not summarize the sender's private reasoning, but should update the shared state with the minimal action-relevant information needed by later agents.

Motivated by this observation, we introduce \textbf{PACT} (\textbf{P}rotocolized \textbf{A}ction-state \textbf{C}ommunication and \textbf{T}ransmission), a minimal and harness-agnostic communication protocol that projects each non-terminal agent's raw output into a compact public action-state message before it is appended to the shared history. Rather than exposing the sender's full generation transcript, PACT retains only the receiver-facing information needed for continuation: the action taken or required next, the task-relevant state, and the resulting artifact to be used downstream. 
Across model scales and MAS settings, PACT consistently reduces token usage while preserving and often improving task performance. We further port PACT to two real-world agentic coding harnesses, OpenHands~\citep{wang2025openhands} and SWE-agent~\citep{yang2024sweagent}.
The same communication rule yields substantial reductions in token usage with minimal performance degradation, indicating PACT is also practical as an inference-time communication protocol for existing agentic applications.

Our contributions are summarized as follows:
\begin{itemize}[leftmargin=*,itemsep=2pt]
  \item 
        Through a systematic analysis of communication strategies in inter-agent message in MAS, we show that no fixed policy is universally optimal, indicating that what agents pass to one another is a central but underexplored design dimension.

  \item 
        We introduce \textbf{PACT}, a training-free and harness-agnostic communication protocol for MAS, defining the boundary between private computation and public communication. 

  \item 
        Experiments on two MAS settings show that PACT substantially improves the performance--cost frontier, reducing token usage by 38.7\% on average across baselines and model scales while preserving or improving task performance.
        On real-world coding harnesses, PACT reduces tokens-per-resolved by 47\% on SWE-agent, demonstrating its practical significance.
\end{itemize}
\section{Related Work}
\label{sec:related}

\paragraph{Multi-agent systems.}
LLM-based multi-agent systems (MAS) coordinate several model instances to
solve tasks beyond a single call. Research has largely explored \emph{who}
talks and \emph{when}: role specialisation and persona
self-collaboration~\citep{wang2023unleashing,li2023camel}, peer debate and
critique~\citep{du2023improving,liang2023encouraging}, general multi-agent
conversation frameworks~\citep{wu2023autogen, chen2024agentverse}, and parallel output aggregation~\citep{wang2024mixture}. These designs differ in roles or orchestration, but uniformly leave the content of each inter-agent message to free-form natural language. We examine through systematic experiments and show that the content one agent pass to the next matters in both performance and cost.

\paragraph{Token consumption in MAS.}
The token cost of multi-agent coordination arises from repeated context reuse:
one agent's output becomes part of another agent's input, and later agents must
reprocess earlier messages. This effect is especially pronounced when agents
forward deliberative content, such as chain-of-thought reasoning~\citep{wei2022chain},
because the same reasoning traces are repeatedly read by downstream agents~\citep{zeng2025s2}.
Prior work studies the \emph{shape} of inter-agent messages and shows that disciplined communication affects accuracy and cost~\citep{zou2025latent, chen2025optima, yu2026learning}. 
While studies pay attention to who should the agent talk to under various topologies~\citep{zhang2025cut, shen2025understanding}, what content should be communicated is largely ignored.
Instead, PACT defines a communication invariant over shared history: only action, grounded state, and reusable result are made public, while private
deliberation remains outside the inter-agent channel.

\paragraph{Agentic harnesses for real-world tasks.}
Beyond research scaffolds, agentic harnesses are increasingly built to solve
real-world problems, most prominently software engineering~\citep{hong2024metagpt, wang2024executable, qian2024chatdev}:
SWE-bench~\citep{jimenez2024swebench} evaluates agents on real GitHub issues,
and OpenHands~\citep{wang2025openhands} and SWE-agent~\citep{yang2024sweagent}
are widely used loops that interleave reasoning, tool calls, and environment
feedback over long trajectories. Because such harnesses accumulate long
tool-augmented histories, the per-turn message and shared-history budget can easily scale up. We show that passing the action-state related information only is enough to reach comparable results while significantly reducing the token usage.


\section{Diagnostic Analysis of Inter-Agent Communication}
\label{sec:prelim}

\begin{figure*}[t]
  \centering
  \includegraphics[width=0.95\textwidth]{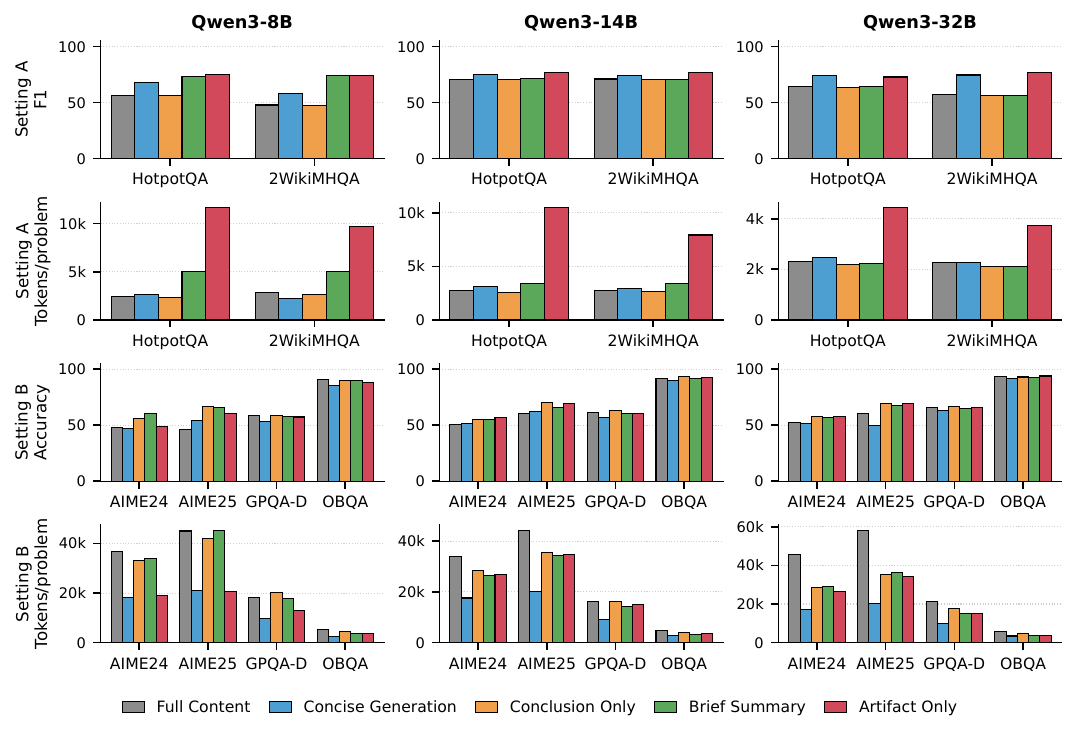}
  \caption{Five inter-agent communication strategies in two MAS settings at three model scales. Top two rows: Setting~A
           interaction (F1 and tokens per problem); bottom two rows: Setting~B
           pipeline (Accuracy and tokens).}
  \label{fig:probe_bars}
  \vspace{-1em}
\end{figure*}

Following the question raised in \S\ref{sec:intro}, we examine how an agent's output should be handled in the inter-agent message, isolating this single choice while holding all other MAS design factors
fixed. We analyze five communication strategies that cover today's common
choices, in two structurally different MAS settings, at various model scales. 

\subsection{Two MAS Settings}
\label{sec:settings}

\paragraph{Setting~A --- Split-Evidence Interaction.}
We use a symmetric two-agent setting where the evidence needed to solve the
task is split across agents. Each agent receives only partial information, so
neither agent can reliably answer the question alone. The agents must exchange
task-relevant evidence over multiple turns and combine their partial views to
produce the final answer. 

\paragraph{Setting~B --- Sequential pipeline.}
In this setting, the task is processed sequentially among the agents, each agent will take the output of previous agents as input and solve the task building on earlier intermediate artifacts.
We adopt a widely studied role-specialised
pipeline~\citep{zou2025latent,yu2026learning} of four agents run in fixed
order---\textbf{Planner}, \textbf{Critic}, \textbf{Refiner}, and
\textbf{Solver}. The Planner produces a plan, the Critic reviews it and
emits a critique, the Refiner returns a revised plan, and the Solver produces
the benchmark answer.

\paragraph{Models and settings.}
All experiments run on Qwen3-8B, Qwen3-14B, and Qwen3-32B~\citep{qwen3team2025}. 
For the split-evidence interaction setting, we evaluate on HotpotQA~\citep{yang2018hotpotqa} and 2WikiMultiHopQA~\citep{ho2020constructing}.
Each answer depends on combining multiple supporting paragraphs and requires evidence exchange. Each question is supportted by ten paragraphs, including two gold supporting paragraphs and eight distractors, and evenly split them between two agents. 
For the sequential pipeline setting, we evaluate on three categories of benchmarks: (i) mathematical reasoning, including AIME2024~\citep{aime2024dataset} and AIME2025~\citep{aime2025dataset}; (ii) scientific reasoning, using GPQA-Diamond~\citep{rein2024gpqa}; and (iii) commonsense question answering, using OpenBookQA~\citep{mihaylov2018openbookqa}.
Other experiment details are deferred to Appendix~\ref{app:details}.

\subsection{Inter-Agent Communication Strategies}
\label{sec:strategies}

We analyze five strategies that cover the common ways a MAS can pass an
agent's reasoning and output on the inter-agent channel.
\textbf{Full Content} appends the agent's freely generated output, including
its internal reasoning trace, to the shared history.
\textbf{Concise Generation} lets the model operate in its native non-thinking mode, which usually produces shorter responses.
\textbf{Conclusion Only} passes the conclusion / final answer portion to the next agent. 
\textbf{Brief Summary} explicitly asks the agent to pass a short free-form summary to the next agent. 
\textbf{Artifact Only} reduces the message to the role artifact alone (e.g., the plan or  critique), without the surrounding action description or supporting state.

\subsection{From Action-Centered Messages to Structured Handoffs}
\label{sec:probe_analysis}

Figure~\ref{fig:probe_bars} shows that no common communication strategy is uniformly satisfactory across the two MAS settings. 
The useful signal in the inter-agent communication is usually action-centered, but it must also expose the state that makes the handoff reliable.
These findings motivate a protocolized handoff that aligns the content of inter-agent messages and makes them compact and stable.

\paragraph{Passing the full content is redundant.}
Full Content forwards the entire free-form output, including the explanations, which is consistently expensive and rarely gives the best performance. 
In the sequential pipeline it has the largest token cost at every model scale, while being clearly behind the best-performing alternatives. 
In the interaction setting it is also not the strongest choice. 
These results show that forwarding the full transcript is an inefficient default: much of what is passed is not the information the receiver needs to act
on.

\paragraph{Being short is not enough.}
Concise Generation reduces token usage by suppressing long reasoning traces, and it performs well in the interaction setting, where the receiver often only needs a surfaced fact. 
However, the same strategy is weak in the role-specialized pipeline, where intermediate agents must produce useful plans, critiques, and refinements before the final solver acts. 
Brief Summary gives a different kind of short message, but its behavior is unstable: it is strong for some 8B interaction and pipeline runs, yet falls behind simpler strategies at larger scales. 
These results suggest that generic conciseness is not a reliable communication rule: a message can be short while still omitting the information needed by the next agent.

\paragraph{Conclusion-only messages are topology-dependent.}
Keeping the visible final output while removing the intermediate traces works relatively well in the sequential pipeline, where the fixed role order already tells each agent how to interpret the previous output. In the interaction setting, however, Conclusion Only is much weaker, because the partner needs not only a local conclusion but also the evidence observed by the sender and the missing information still needed for the answer. 
Thus, removing thinking processes is helpful, but keeping only the conclusion is not a general solution: it depends on whether the surrounding MAS topology already supplies the missing action and state information.

\begin{figure}[t]
  \centering
  \includegraphics[width=0.78\columnwidth]{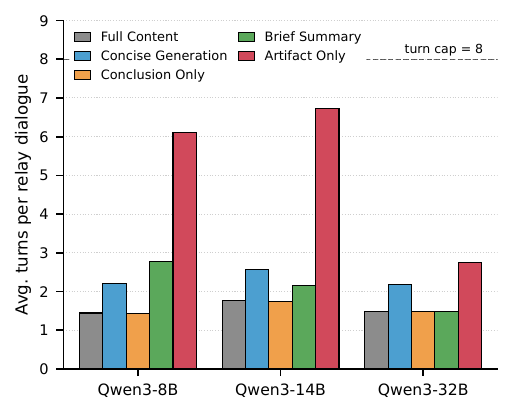}
  \caption{Average agent turns
           per interaction dialogue: \emph{Artifact Only} runs
           more turns than other strategies,
           driving its $\sim$3$\times$ token cost.}
  \label{fig:artifact_turns}
  \vspace{-10pt}
\end{figure}

\paragraph{Artifact-only messages identify the right content but not the right protocol.}
Artifact Only provides the clearest signal about what should be communicated.
It often achieves the best in the interaction setting and is competitive in the
sequential pipeline, showing that the receiver benefits most from the sender's
action-related work product rather than from the full generation trace. 
However, Artifact Only is not yet a communication protocol. In the interaction setting it can achieve high accuracy only with very large messages. As shown in Figure~\ref{fig:artifact_turns}, Artifact Only runs more turns than other strategies,  indicating that suppressing all preamble also suppresses the closure cues agents use to stop. 
In the pipeline, it is efficient and competitive at larger scales, but weaker for the smaller model, suggesting that an artifact alone may omit useful state or handoff information when the sender is less capable. 
The artifact is therefore the right starting point, but it must be made structured and compact.





\section{PACT: Protocolized Action-state Communication and Transmission}
\label{sec:method}

\begin{figure}[t]
  \centering
  \includegraphics[width=0.95\columnwidth]{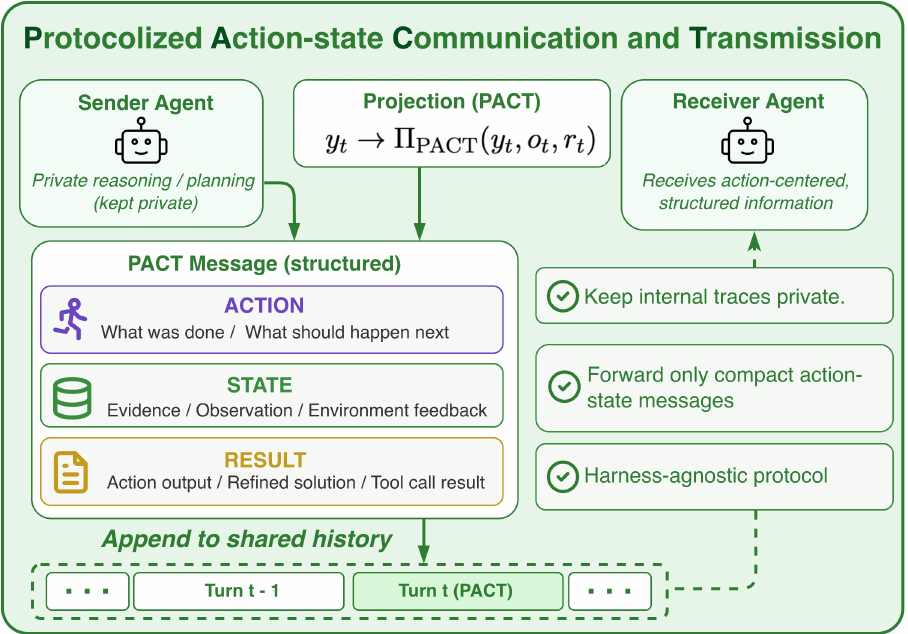}
  \caption{Overview of PACT
  }
  \label{fig:method}
  \vspace{-10pt}
\end{figure}

The analysis result in \S\ref{sec:prelim} suggests that the central challenge in
inter-agent communication is not merely message length, but the type of
information that is preserved for downstream agents. Rather than exposing the
sender's full generation transcript, an inter-agent message should communicate
the sender's action-state: the action taken, the task-relevant state or
evidence supporting it, and the result that the next agent needs in order to
continue. Building on this observation, we propose \textbf{Protocolized Action-state Communication and Transmission}
(\textbf{PACT}), shown in Figure~\ref{fig:method}, a communication protocol that
restricts shared inter-agent history to compact action-state messages while
excluding intermediate process-level content.

\subsection{General setup}
\label{sec:pact_setup}

Consider a multi-agent system with a shared history $H_t$ at turn $t$. An agent
with role or action $r_t$ receives its local observation $o_t$ and the shared
history $H_t$, then produces a raw output $y_t$. In a standard MAS, this raw
output is directly appended to the shared history:
\begin{equation}
    H_{t+1}^{\mathrm{std}} = H_t^{\mathrm{std}} \oplus y_t ,
\end{equation}
where $\oplus$ denotes appending a message. This update rule is simple, but it
makes every part of $y_t$ public, including reasoning traces, thought explanations, repeated statements, and the final artifact. These elements are all retained and
re-read by later agents.

The idea of PACT is simple: before an agent output becomes shared history, we
project it into a compact public message space. Formally, PACT replaces the raw
message $y_t$
and updates the history as
\begin{equation}
    H_{t+1}^{\mathrm{PACT}}
    =
    H_t^{\mathrm{PACT}} \oplus \Pi_{\mathrm{PACT}}(y_t, o_t, r_t) .
\end{equation}
Here $\Pi_{\mathrm{PACT}}$ is a sender-side projection that determines what
information is allowed to enter the inter-agent channel. 

\subsection{Action-state message space}
\label{sec:pact_message_space}

PACT represents each public inter-agent message as an action-state record with
three receiver-facing fields: \textsc{Action}, \textsc{State}, and
\textsc{Result}. \textsc{Action} states what the sender has done, or what the
next agent should do; this makes the handoff explicit rather than leaving the
receiver to infer the sender's intent. \textsc{State} records the evidence,
observation, environment feedback, or tool result that grounds the message; this
allows the receiver to consume the message as a grounded public state rather
than an unsupported conclusion. \textsc{Result} contains the action output
itself, such as a retrieved fact, a refined solution, or a tool-call result. Together, these fields define the information that is allowed
to enter the shared inter-agent channel.

This message space also specifies the boundary of PACT.
Each agent may still reason, self-check, or explore alternatives as needed to produce its output. The constraint applies only to what becomes part of the public inter-agent message space. Under PACT, a message retains only the information that is relevant to the action-state record: what action was taken, what state or evidence grounds it, and what result should be passed downstream. 
Intermediate process content such as hidden reasoning traces or redundant restatements is excluded before the message is appended to the shared history. 
In this way, PACT separates private computation from public communication
, preventing intermediate deliberation from compounding across later context windows.

\subsection{Protocol properties}
\label{sec:pact_properties}

Since PACT is intended to be a protocol over the inter-agent channel, rather
than a template tied to a particular MAS scaffold, it has the following
properties. First, PACT does not constrain the agent's private computation. The
agent can still perform internal reasoning and planning in its original manner; PACT
only controls which part of the generated output is retained as public
communication. This differs from methods that suppress reasoning or alter the
agent's internal solving process and preserves the capability of the agent. 
Second, PACT is orthogonal to standard MAS
design choices. It does not introduce new agents, change the turn schedule, train a compressor, or assume a particular role structure.
The same action-state interface can be serialized differently across settings:
a split-evidence interaction may write the action, state, and result explicitly, while a
role-specialized pipeline may encode part of the action through the role
schedule and retain a more compact role artifact. 
This protocol works for controlled MAS scaffolds and complex tool-using harnesses like agentic coding systems.



\section{Experiments}
\label{sec:experiments}

\subsection{Experiment setup}
\label{sec:exp_setup}

We evaluate on the same two MAS settings introduced in
\S\ref{sec:prelim}: the two-agent split-evidence interaction (Setting~A) and the sequential pipeline (Setting~B), over the same six benchmarks.
All experiments use Qwen3-8B, Qwen3-14B, and Qwen3-32B~\citep{qwen3team2025}.
We compare PACT against three representative MAS communication paradigms.
\textbf{Chain of Agents (CoA)}~\citep{zhang2024chain} chains worker agents
in sequence, with each agent reading only the most recent peer message
rather than the full inter-agent history. \textbf{Text-based MAS
(TextMAS)}~\citep{zou2025latent} uses role-specialised agents that
collaborate through unconstrained natural-language messages and retain the
full message history in the shared channel. \textbf{Multi-Agent
Debate}~\citep{du2023improving} runs multiple agents that propose initial
answers and then debate over several rounds, converging on a final answer
by majority vote. 
We report token-overlap F1 for the interaction setting and exact-match accuracy for the pipeline, alongside the average total tokens per problem.
Implementation details for each baseline are given in Appendix~\ref{app:details}.

\begin{table}[t]
\footnotesize
\centering
\setlength{\tabcolsep}{3pt}
\caption{\textbf{PACT vs.\ baselines on the two-agent split-evidence interaction
         (Setting~A).} Tok $=$ avg total tokens per problem. Best F1 and lowest tokens per benchmark
         within each model in \textbf{bold}.}
\label{tab:relay_main}
\begin{tabular}{l rr rr}
\toprule
& \multicolumn{2}{c}{\textbf{HotpotQA}} & \multicolumn{2}{c}{\textbf{2WikiMHQA}} \\
\cmidrule(lr){2-3}\cmidrule(lr){4-5}
Method & F1 & Tok & F1 & Tok \\
\midrule
\multicolumn{5}{l}{\rlap{\textit{Qwen3-8B}}} \\
CoA      & 68.6 & 7,281 & 65.6 & 6,744 \\
TextMAS  & 59.5 & 9,384  & 58.0 & 7,753 \\
Multi-Agent Debate   & 69.6   & 13,520     & 68.2   & 10,669     \\
\rowcolor{lightblue}
\textbf{PACT (ours)} & \textbf{69.9} & \textbf{6,704} & \textbf{68.7} & \textbf{6,100} \\
\midrule
\multicolumn{5}{l}{\rlap{\textit{Qwen3-14B}}} \\
CoA      & 54.8 & 6,782 & 55.2 & 6,452 \\
TextMAS  & 45.8 & 8,652  & 41.8 & 7,160 \\
Multi-Agent Debate   & 48.4   & 12,133     & 36.7   & 9,561     \\
\rowcolor{lightblue}
\textbf{PACT (ours)} & \textbf{56.8} & \textbf{5,996} & \textbf{59.2} & \textbf{5,130} \\
\midrule
\multicolumn{5}{l}{\rlap{\textit{Qwen3-32B}}} \\
CoA      & 54.2 & 5,871 & 56.3 & 5,339 \\
TextMAS  & 50.1 & 8,857  & 46.4 & 7,300 \\
Multi-Agent Debate   & 60.3   & 12,258     & 58.7   & 9,467     \\
\rowcolor{lightblue}
\textbf{PACT (ours)} & \textbf{61.5} & \textbf{4,821} & \textbf{61.3} & \textbf{4,039} \\
\bottomrule
\end{tabular}
\end{table}

\begin{table*}[t]
\footnotesize
\centering
\setlength{\tabcolsep}{3pt}
\caption{\textbf{PACT vs.\ baselines on the four-agent sequential pipeline (Setting~B).}
         Tok $=$ avg total tokens per problem. PACT row
         shaded; best accuracy and lowest tokens per benchmark within each
         model in \textbf{bold}.
         }
\label{tab:as_pipeline}
\begin{tabular}{l *{5}{cc}}
\toprule
& \multicolumn{2}{c}{AIME24} & \multicolumn{2}{c}{AIME25} & \multicolumn{2}{c}{GPQA-D} & \multicolumn{2}{c}{OBQA} & \multicolumn{2}{c}{\textbf{Mean}} \\
\cmidrule(lr){2-3}\cmidrule(lr){4-5}\cmidrule(lr){6-7}\cmidrule(lr){8-9}\cmidrule(lr){10-11}
Method & Acc & Tok & Acc & Tok & Acc & Tok & Acc & Tok & Acc & Tok \\
\midrule
\multicolumn{11}{l}{\rlap{\textit{Qwen3-8B}}} \\
CoA      & 55.0 & 38,072 & 63.7 & 47,279 & 59.5 & 22,563 & 90.4 & 4,851 & 67.2 & 28,191 \\
TextMAS  & 48.3 & 36,911 & 45.8 & 45,004 & 59.0 & 20,357 & 91.1 & 5,263 & 61.1 & 26,884 \\
Multi-Agent Debate   & 56.4 & 140,863 & \textbf{70.8} & 169,785 & 57.7 & 85,351 & 92.0 & 32,784 & 69.2 & 107,196 \\
\rowcolor{lightblue}
\textbf{PACT (ours)} & \textbf{57.5} & \textbf{31,414} & 67.5 & \textbf{41,071} & \textbf{60.4} & \textbf{19,960} & \textbf{92.5} & \textbf{4,304} & \textbf{69.5} & \textbf{24,187} \\
\midrule
\multicolumn{11}{l}{\rlap{\textit{Qwen3-14B}}} \\
CoA      & 54.9 & 28,368 & 64.5 & 35,807 & 61.6 & 18,789 & 93.9 & 4,533 & 68.7 & 21,874 \\
TextMAS  & 50.8 & 34,066 & 60.8 & 44,111 & 61.0 & 16,366 & 92.0 & 4,675 & 66.2 & 24,805 \\
Multi-Agent Debate   & 55.2 & 138,081 & \textbf{74.3} & 156,895 & 61.8 & 81,878 & 93.2 & 30,704 & 71.1 & 101,890 \\
\rowcolor{lightblue}
\textbf{PACT (ours)} & \textbf{55.4} & \textbf{26,994} & 71.6 & \textbf{32,548} & \textbf{62.5} & \textbf{15,935} & \textbf{95.2} & \textbf{3,559} & \textbf{71.2} & \textbf{19,759} \\
\midrule
\multicolumn{11}{l}{\rlap{\textit{Qwen3-32B}}} \\
CoA      & 57.3 & 29,593 & 71.2 & 35,818 & 65.7 & 18,535 & 93.1 & 3,811 & 71.8 & 21,939 \\
TextMAS  & 52.1 & 45,867 & 60.4 & 57,984 & 65.7 & 21,428 & 93.8 & 5,770 & 68.0 & 32,762 \\
Multi-Agent Debate   & 58.4 & 134,679 & \textbf{75.8} & 157,744 & 66.4 & 77,888 & 93.7 & 28,394 & 73.6 & 99,676 \\
\rowcolor{lightblue}
\textbf{PACT (ours)} & \textbf{59.2} & \textbf{25,574} & 72.7 & \textbf{30,970} & \textbf{68.5} & \textbf{16,271} & \textbf{94.4} & \textbf{3,517} & \textbf{73.7} & \textbf{19,083} \\
\bottomrule
\vspace{-20pt}
\end{tabular}
\end{table*}

\subsection{Main Results}
\label{sec:results}

Tables~\ref{tab:relay_main} and~\ref{tab:as_pipeline}  show that PACT shifts the performance--cost frontier across both evaluated MAS settings, demonstrating the effectiveness of passing the action-state information.

\paragraph{Grounded handoffs outperform full-history communication.}
In the split-evidence interaction setting, a successful message must do more than state a local conclusion: it must surface the relevant evidence and clarify what information is still needed from the other agent. 
This requirement exposes the weakness of unconstrained communication. 
For example, TextMAS retains the full conversation history, but its free-form generations introduce redundant reasoning and force later turns to re-read unnecessary content. 
As shown in Table~\ref{tab:relay_main}, it consumes more tokens than PACT but still performs worse on both benchmarks, indicating that preserving more text does not necessarily provide a more useful handoff. 
In contrast, as shown in the example in Figure~\ref{fig:casestudy}, PACT fits the interaction structure by using an explicit action-state format, turning each turn into a grounded handoff rather than a free-form explanation or an unsupported answer fragment. 
This allows PACT to achieve the best performance while using fewer tokens.

\paragraph{Compact artifacts outperform extended deliberation.}
In the sequential pipeline setting, the communication requirement is different: each downstream agent needs the artifact-related information to update its own planning. 
This exposes a different limitation of unrestricted multi-agent interaction. 
Although Multi-Agent Debate can improve reasoning through repeated answer-level deliberation, its communication is not organized around the intermediate artifacts needed, causing substantial token overhead. 
As shown in Table~\ref{tab:as_pipeline}, PACT achieves comparable or better performance on most benchmarks while using only 19\%--23\% of the tokens required by Multi-Agent Debate. 
This suggests that the pipeline benefits less from extended deliberation than from passing compact action-related artifacts that downstream agents can directly use.


\paragraph{Stronger models make better use of compact communication.}
The scale trend further supports the role of sparse public communication. 
Under PACT, stronger models require less explicit inter-agent communication while achieving better task performance. 
In the sequential pipeline, PACT's average token usage decreases by 21.1\% from Qwen3-8B to Qwen3-32B, while increasing the mean accuracy by 4.2 points. 
A similar reduction appears in the interaction setting, where PACT's token usage decreases as the model scale increases on both benchmarks. 
This suggests that stronger agents can better exploit compact evidence and refined artifacts, without requiring the sender's full reasoning process to be repeatedly exposed in the shared history.
However, under baseline settings, larger models continue to spend tokens producing and consuming redundant public content.
Thus, pure model scaling cannot fix communication inefficiency: lacking a protocol, powerful models still waste tokens on redundant public messages, while PACT enables more private computation and keeps agent communication concise.

\subsection{Ablation Study}
\label{sec:ablation}

Table~\ref{tab:ablation} shows how each field in PACT contributes to effective communication. 
Removing \textsc{Action} reduces F1 from 69.9 to 64.9, showing that the receiver needs an explicit signal about what information is being provided or requested. 
Removing \textsc{State} also hurts performance, which indicates that the action result is less useful when it is not grounded in the sender's observed evidence. 
The largest degradation appears when both action and state are removed and only \textsc{Result} is forwarded: F1 falls to 64.3 and token usage increases by 12.9\%. 
This suggests that an unsupported result creates ambiguity for the receiver, leading to both worse task performance and less efficient interaction. 
Overall, the ablation confirms that PACT's advantage comes from the complete action-state handoff: the action clarifies how the message should be used, the state grounds the message in evidence, and the result carries the factual contribution.

\begin{table}[t]
\footnotesize
\centering
\setlength{\tabcolsep}{4pt}
\caption{PACT field-ablation on HotpotQA with Qwen3-8B. A~=~Action, S~=~State, R~=~Result.}
\label{tab:ablation}
\begin{tabular}{l r r r r}
\toprule
 & \textbf{F1} & \textbf{Tok} & \textbf{$\Delta$F1} & \textbf{$\Delta$Tok} \\
\midrule
\textbf{PACT (A+S+R)}                              & 69.9 & 6,704 & --- & --- \\
\quad w/o A \ (S+R)                                & 64.9 & 6,826 & $-5.0$ & $+1.8\%$ \\
\quad w/o S \ (A+R)                                & 65.2 & 6,741 & $-4.7$ & $+0.6\%$ \\
\quad w/o A, S \ (R only)                          & 64.3 & 7,571 & $-5.6$ & $+12.9\%$ \\
\bottomrule
\end{tabular}
\end{table}

\section{PACT on Agentic Coding Harnesses}
\label{sec:harnesses}

The settings so far use predefined MAS scaffolds. We now test whether the
communication-content rule transfers to external production agentic coding harnesses it was not designed for, such as \textbf{OpenHands}
(CodeActAgent)~\citep{wang2025openhands} and \textbf{SWE-agent}~\citep{yang2024sweagent}, on \textbf{SWE-bench Verified}~\citep{jimenez2024swebench}, which solves real-world GitHub issues in real environments.

\subsection{Porting PACT as a proxy hook}
\label{sec:harness_hook}

PACT is implemented as a proxy hook that updates the
public messages passed between agent turns in-flight. The hook has two
components. First, each turn must emit, before its tool call, a structured
\texttt{<summary>} block containing \textit{Action Required},
\textit{Observed State}, and \textit{Planned Effect}---the
\S\ref{sec:method} schema rephrased with forward-looking coding semantics.
Second, before each turn the proxy rewrites prior assistant messages to
keep only the \texttt{<summary>} block and the tool calls, removing the
intermediate process and any free-form prose; tool results are left intact,
so the inter-turn channel reduces to the information of action-state summary, tool calling and tool results.


\subsection{Results on coding harnesses}
\label{sec:harness_results}

As shown in Table~\ref{tab:harness}, PACT shifts the efficiency frontier in both coding harnesses. On \textbf{OpenHands}, PACT improves both
effectiveness and efficiency: the SWE-bench Verified resolve rate increases
from $19.40\%$ to $23.00\%$ ($+18$ resolved instances), while
tokens-per-resolved decreases by $10.3\%$ and average completion tokens per
call decrease by $5.3\%$. On \textbf{SWE-agent}, PACT produces an even larger
efficiency gain, reducing input tokens from $314.6\,\text{M}$ to
$156.0\,\text{M}$ ($-50.4\%$), which is the dominant cost in this
long-context loop with only a small resolve-rate change, while tokens-per-resolved drops by approximately
$47\%$. Thus, the efficiency
gain is consistent across both production coding harnesses.

This indicates that PACT is not tied to the controlled MAS
scaffolds used in earlier sections. The same action-state principle can be
implemented as a lightweight proxy-level intervention and still improves the
efficiency of external coding agents. It improves OpenHands, where verbose assistant turns are repeatedly
carried through the interaction history, and remains approximately neutral on
SWE-agent, where the main gain comes from reducing accumulated input context.
In both cases, PACT reduces the cost per solved instance, suggesting
that sparse public communication is a portable efficiency mechanism even when
the underlying agent loop and tool interface differ.
\begin{table}[t]
\small
\centering
\setlength{\tabcolsep}{4pt}
\caption{\textbf{PACT ported to production agentic coding harnesses}
         on SWE-bench Verified, Qwen3-14B.}
\label{tab:harness}
\begin{tabular}{lrrr}
\multicolumn{4}{l}{\emph{OpenHands (CodeActAgent)}} \\
\toprule
Metric & Baseline & \textbf{PACT} & $\Delta$ \\
\midrule
Resolved            & 97/500 & \textbf{115/500} & \textbf{+18} \\
Resolved \%         & 19.40 & \textbf{23.00} & \textbf{+3.60 pp} \\
Avg compl./call     & 870.6 & \textbf{824.2} & \textbf{$-$5.3\%} \\
Tokens / resolved   & 3.82\,M & \textbf{3.43\,M} & \textbf{$-$10.3\%} \\
\bottomrule
\end{tabular}

\vspace{4pt}

\begin{tabular}{lrrr}
\multicolumn{4}{l}{\emph{SWE-agent (default single-agent)}} \\
\toprule
Metric & Baseline & \textbf{PACT} & $\Delta$ \\
\midrule
Resolved            & 128/500 & 121/500 & $-7$ \\
Resolved \%         & 25.6 & 24.2 & $-1.4$ pp \\
Input tokens        & 314.6\,M & \textbf{156.0\,M} & \textbf{$-$50.4\%} \\
LLM calls           & 21,398 & \textbf{20,623} & \textbf{$-$3.6\%} \\
Tokens / resolved   & 2.46\,M & \textbf{1.30\,M} & \textbf{$-$47\%} \\
\bottomrule
\end{tabular}
\end{table}

\section{Conclusion}
\label{sec:conclusion}


This paper studies what agents should pass to one another in MAS. Through diagnostic analysis across five common strategies on two MAS settings, we show that no common communication strategy is uniformly optimal, and the useful content in an inter-agent message is usually action-centered. Motivated by this, we propose \textbf{PACT} (\textbf{P}rotocolized \textbf{A}ction-state \textbf{C}ommunication and \textbf{T}ransmission), a communication protocol that restricts shared history to compact action-state messages while excluding intermediate content. 
PACT only updates the content in the public messages without affecting the capability of the agent model. 
Across controlled MAS settings and production coding harnesses, PACT improves the performance--cost trade-off by reducing redundant context while preserving or improving task performance. 
These results suggest that disciplining what enters the shared inter-agent channel is itself an effective mechanism for reducing redundant context growth in inter-agent communication, and we should not merely leave it as unconstrained natural-language outputs.
\section*{Limitations}


PACT is designed for MAS settings where shared conversational history is a major source of token cost, its benefits on systems with short interactions or architectures that do not repeatedly expose prior agent outputs are not fully explored. Our experiments cover two controlled MAS topologies and two agentic coding harnesses, but they do not exhaust all forms of multi-agent collaboration, such as open-ended debate, tool-heavy planning, or dynamically routed agent networks.


\bibliography{custom}
\clearpage
\appendix
\section*{Appendix}
\section{Experimental Details}
\label{app:details}

This section collects the experimental settings for both the diagnostic
analysis (\S\ref{sec:prelim}) and the PACT experiments
(\S\ref{sec:experiments},~\S\ref{sec:harnesses}). Unless noted otherwise, all
runs use vLLM with sampling temperature $0.6$ and top-$p$ $0.95$.

\paragraph{Models.}  Our evaluation includes models from the Qwen3 family~\citep{qwen3team2025} (8B, 14B and 32B in scale), allowing us to
assess performance across both mid-scale and large-scale reasoning models.

\paragraph{Benchmarks.} For the split-evidence interaction, we evaluate on HotpotQA~\citep{yang2018hotpotqa} and 2WikiMultiHopQA~\citep{ho2020constructing}.
For the sequential pipeline setting, we evaluate on three categories of benchmarks: (i) mathematical reasoning, including AIME2024~\citep{aime2024dataset} and AIME2025~\citep{aime2025dataset}; (ii) scientific reasoning, using GPQA-Diamond~\citep{rein2024gpqa}; and (iii) commonsense question answering, using OpenBookQA~\citep{mihaylov2018openbookqa}.
We report AIME24/25 and GPQA-Diamond as avg@8 where we average over different seeds to resolve sampling noise; single-seed benchmarks use seed $42$.

\paragraph{Diagnostic analysis (\S\ref{sec:prelim}).}
The diagnostic analyses evaluate five common inter-agent communication
strategies in both MAS settings: \emph{Full Content}, \emph{Concise Generation},
\emph{Conclusion Only}, \emph{Brief Summary}, and \emph{Artifact Only}. In the
two-agent split-evidence interaction, we use a 5--5 context split, where each agent
receives one gold supporting paragraph and four distractors. Agents communicate
for at most $8$ turns with early exit, and we set \texttt{max\_new\_tokens}
to $8{,}192$. 
The sequential pipeline uses the same five communication strategies under the fixed Planner--Critic--Refiner--Solver order, with benchmark-specific settings
reported below.

\paragraph{PACT experiments.}
The four-agent Planner$\rightarrow$Critic$\rightarrow$Refiner$\rightarrow$Solver
pipeline (\S\ref{sec:prelim} Setting~B, \S\ref{sec:experiments}) and the
two-agent split-evidence interaction (\S\ref{sec:experiments}) reuse the corresponding
analysis settings. The split-evidence interaction uses $4$ alternating
turns, \texttt{max\_new\_tokens}$=4{,}096$ per turn, 5--5 split.
Following~\citet{yu2026learning}, the per-benchmark settings are given in
Table~\ref{tab:app_settings}.

\paragraph{Baselines (\S\ref{sec:experiments}).}
The three MAS baselines are instantiated as follows. All baselines share the
decoding settings, per-benchmark token budgets, and dataset splits described
above; only the inter-agent communication protocol differs.

\noindent\textbf{Chain of Agents (CoA).}~\citep{zhang2024chain}
We use the same agent counts as the corresponding MAS settings ($2$ workers
for the interaction setting, $4$ for the pipeline). Each worker reads only the
\emph{single} most recent message from its predecessor---there is no rolling
shared history, so the inter-agent channel carries only $1$ round of context
between consecutive workers.

\noindent\textbf{Text-based MAS (TextMAS).}~\citep{zou2025latent}
Role-specialised agents collaborating via free-form natural-language
messages, with the full message history retained in the shared channel and
no constraint on the message format. We instantiate it under the same MAS
configurations: two symmetric agents with $4$ alternating turns for
Setting~A, and the four-agent Planner$\rightarrow$Critic$\rightarrow$Refiner%
$\rightarrow$Solver schedule for Setting~B. 

\noindent\textbf{Multi-Agent Debate.}~\citep{du2023improving}
$4$ agents independently propose initial answers and then debate over $3$
rounds, each round revising its answer conditioned on the other agents'
previous-round answers; the final answer is the majority vote across agents
after the last round. The same configuration is used in both settings.

\begin{table}[h]
\centering
\small
\setlength{\tabcolsep}{6pt}
\caption{Per-benchmark evaluation settings for the sequential pipeline and the
         split-evidence interaction.}
\label{tab:app_settings}
\begin{tabular}{lcc}
\toprule
Benchmark & $n$ & \texttt{max\_new\_tokens} \\
\midrule
AIME 2024 / 2025 & $30$ each & $32{,}768$ \\
GPQA-Diamond     & $198$     & $8{,}192$  \\
OpenBookQA       & $1{,}000$ & $2{,}048$  \\
HotpotQA         & $7{,}405$ & $4{,}096$  \\
2WikiMultiHopQA  & $12{,}576$& $4{,}096$  \\
\bottomrule
\end{tabular}
\end{table}

\paragraph{Agentic coding harnesses (\S\ref{sec:harnesses}).}
OpenHands (CodeActAgent) and SWE-agent are evaluated on SWE-bench Verified with Qwen3-14B. PACT is applied as a LiteLLM proxy hook that rewrites prompts and parses responses in flight, with no SDK or docker-image changes.

\section{Split-Evidence Interaction Case Study: How PACT Changes the Message}
\label{app:casestudy}

Figure~\ref{fig:casestudy} contrasts the same interaction turn with and without
PACT on a representative HotpotQA bridge question (\emph{``Which magazine
was started first, Arthur's Magazine or First for Women?''}). Without
PACT---essentially the TextMAS baseline---the agent forwards its full
\texttt{<think>} deliberation and hedged prose, so the partner re-reads
hundreds of tokens of reasoning about the agent's own paragraph. PACT keeps
exactly the surfaced fact and the precise request: the
\textasciitilde$10\times$ shorter message carries the same actionable
content, which is why downstream turns stay accurate while the shared
history stops compounding.

\begin{figure}[t]
\centering
\begin{tcolorbox}[enhanced, colback=white, colframe=black!55, boxrule=0.4pt,
  fonttitle=\bfseries\footnotesize, fontupper=\footnotesize,
  left=3pt, right=3pt, top=2pt, bottom=2pt,
  title={Without PACT (free-form, \texttt{<think>} kept) --- \textasciitilde480 tokens}]
\texttt{\textless think\textgreater} The question asks which magazine started
first. My paragraph is about \emph{Arthur's Magazine}; it says it was an
American literary periodical published in Philadelphia, established in 1844. I
have nothing on \emph{First for Women}, so I should report my finding and ask
the partner \emph{(\dots\ \textasciitilde420 tokens of deliberation \dots)}
\texttt{\textless/think\textgreater}\\[2pt]
Hi partner! I checked my documents. I have a paragraph on Arthur's Magazine,
and it looks like it was established around 1844 in Philadelphia. I'm not
fully certain, but could you see whether you have anything on First for Women
and when it started, so we can decide which came first?
\end{tcolorbox}
\begin{tcolorbox}[enhanced, colback=asyellow, colframe=black!55, boxrule=0.4pt,
  fonttitle=\bfseries\footnotesize, fontupper=\footnotesize,
  left=3pt, right=3pt, top=2pt, bottom=2pt,
  title={With PACT (three-field) --- \textasciitilde40 tokens}]
\textbf{Action Required:} the founding year of \emph{First for Women}.\\
\textbf{Environment State:} ``Arthur's Magazine was an American literary
periodical first published in 1844.''\\
\textbf{Action Result:} Arthur's Magazine was founded in 1844; need First for
Women's date to compare.
\end{tcolorbox}
\caption{Illustrative interaction turn with and without PACT (HotpotQA bridge
         question). PACT drops the \texttt{<think>} trace and hedged prose,
         keeping only the surfaced fact and the explicit request---an
         \textasciitilde$10\times$ shorter message with the same actionable
         content. Long spans are elided as \emph{(\dots\ tokens \dots)}.}
\label{fig:casestudy}
\end{figure}

\end{document}